%% file: main.tex
\documentclass[10pt,twocolumn,letterpaper]{article}

\usepackage[pagenumbers]{iccv} 


\usepackage{booktabs}
\usepackage{multirow}
\usepackage{algorithm}
\usepackage{algpseudocode}
\usepackage{svg}

\definecolor{iccvblue}{rgb}{0.21,0.49,0.74}
\usepackage[pagebackref,breaklinks,colorlinks,allcolors=iccvblue]{hyperref}

\def\paperID{8094} 
\def\confName{ICCV}
\def\confYear{2025}

\title{FastCAR: Fast Classification And Regression 
\protect\\ for Task Consolidation in Multi-Task Learning 
\protect\\ to Model a Continuous Property Variable of Detected Object Class}

\author{Anoop Kini\thanks{Corresponding author} \quad Andreas Jansche \quad  Timo Bernthaler \quad Gerhard Schneider\\
Hochschule Aalen, Aalen, Germany\\
{\tt\small \{anoop.kini, andreas.jansche, timo.bernthaler, gerhard.schneider\}@hs-aalen.de}
}

\begin{document}
\maketitle
\input{sec/0_abstract}    
\input{sec/1_intro}

\input{sec/2_formatting}
\input{sec/3_finalcopy}
{
    \small
    \bibliographystyle{ieeenat_fullname}
    \bibliography{main}
}

\input{sec/X_suppl}

\end{document}

%% file: sec/0_abstract.tex
\begin{abstract}

FastCAR is a novel task consolidation approach in Multi-Task Learning (MTL) for a classification and a regression task, despite the non-triviality of task heterogeneity with only a subtle correlation. The approach addresses the classification of a detected object (occupying the entire image frame) and regression for modeling a continuous property variable (for instances of an object class), a crucial use case in science and engineering. FastCAR involves a label transformation approach that is amenable for use with only a single-task regression network architecture. FastCAR outperforms traditional MTL model families, parametrized in the landscape of architecture and loss weighting schemes, when learning both tasks are collectively considered (classification accuracy of 99.54\%, regression mean absolute percentage error of 2.4\%). The experiments performed used ``Advanced Steel Property Dataset'' contributed by us \url{https://github.com/fastcandr/Advanced-Steel-Property-Dataset}. The dataset comprises 4536 images of 224x224 pixels, annotated with discrete object classes and its hardness property that can take continuous values. Our proposed FastCAR approach for task consolidation achieves training time efficiency (2.52x quicker) and reduced inference latency (55\% faster) than benchmark MTL networks.

\end{abstract}

%% file: sec/1_intro.tex
\begin{figure}[t]
  \centering
   \includegraphics[width=0.97\linewidth]{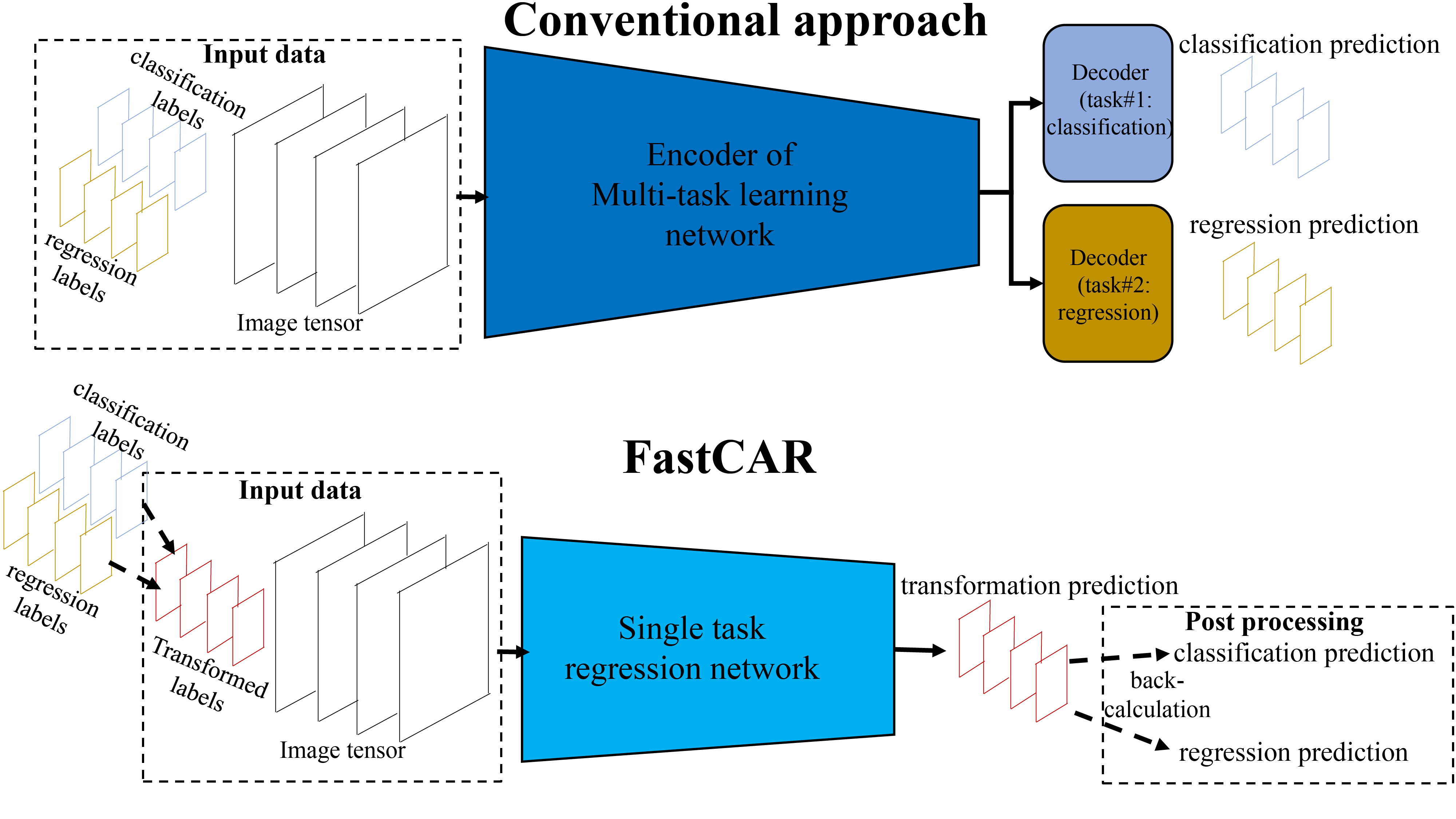}

   \caption{This is a schematic representation of the FastCAR approach for the classification of a detected object instance and the regression for its property prediction that can take continuous values (heterogeneous tasks with subtle correlation) via label transformation and only a single task regression network. Its comparison with conventional multi-task learning is shown.}
   \label{fig:FastCAR_approach}
\end{figure}

\section{Introduction}
\label{sec:intro}

Multi-Task Learning (MTL) networks are widely used to optimize multiple objective functions \cite{sener2018multi, di2023multi} simultaneously, using just one model. Apart from achieving reduced latency benefits \cite{guo2020learning, sagduyu2024low}, MTL can prospectively learn more robust or universal representations \cite{zhang2021survey} than those with several single-task models. In this way, MTL supports better generalization across all tasks via shared feature representation \cite{zhang2021survey, liu2019end} while striving for low latency. MTL networks are widely used in computer vision for object detection and scene understanding, among many other applications that involve a combination of a classification and a regression task, referred to as heterogeneous tasks \cite{zhang2021survey, yang2009heterogeneous}.

 For achieving performance with MTL networks, including for heterogeneous tasks, task-shared (generalizable) feature representations and task-specific feature representations are both critical \cite{liu2019end}. The two requirements broadly translate to two popular experimental knobs in MTL; network architecture and weighting scheme for loss functions, which collectively govern the performance of MTL networks \cite{liu2019end, misra2016cross}. The constituting individual tasks must not hurt the performance of one another, often referred to as negative transfer of knowledge  \cite{guo2020learning, tang2020progressive, standley2020tasks}. The resolution could frequently demand effective task grouping.

In object detection-localization, task grouping of classification and regression tasks (heterogeneous tasks) are frequently implemented within an MTL network by utilizing a common backbone, like in the YOLO algorithm family \cite{terven2023comprehensive}. A possibility of a strong correlation \cite{li2020generalized} between classification and regression tasks (heterogeneous tasks) allows such task grouping. 

Task grouping of heterogeneous tasks with only subtle correlation is non-trivial and can be very challenging \cite{zhang2014facial} due to differences in learning difficulties and convergence rates between tasks. It often requires new approaches in the landscape of architecture and weighting schemes for loss functions. Zhang \etal developed a new architecture called ``Tasks-Constrained Deep Convolutional Network'' \cite{zhang2014facial} to jointly optimize facial landmark detection with subtly correlated tasks, involving classification and regression. Ott \etal \cite{ott2022joint} navigated the challenge of heterogeneous and subtly correlated tasks, for classification (of multivariate time series) and regression (of handwritten trajectory), by developing a new combination of architecture and weighting schemes for loss functions.

We present a novel task grouping approach for heterogeneous tasks with subtle correlation, specifically for the classification of a detected object and modeling its regression-based property variable, called FastCAR (Fast Classification And Regression) as shown in \cref{fig:FastCAR_approach}. We define ``task consolidation'' as a specific case of task grouping, where both tasks accomplish learning via a single task network, exemplified by FastCAR that leverages a novel label transformation strategy.   

The use case involves the classification of a detected object instance (occupies the entire image frame) and a regression task to model a continuous variable, characteristic of an object class across its instances, referred to as ``property modeling''. Such use cases are of paramount relevance across scientific research and engineering disciplines. On our newly contributed dataset called ``Advanced Steel Property Dataset'', FastCAR handles task heterogeneity with only subtle correlation. The simplicity of the architecture and the labeling transformation collectively achieve training time efficiency and reduced inference latency in the domain of MTL. 

Our contribution is three-fold:
\begin{itemize}
\item We investigate and contribute an approach towards task consolidation (task grouping) of heterogeneous and subtly correlated tasks in multi-task deep learning, involving a classification and a regression task, via label transformation.

\item In this context, we identify a use case for object classification and its property modeling via regression. The generic nature of problem formulation, involving an object instance occupying the entire image frame and its property prediction, is significantly relevant for a wide range of science and engineering disciplines.  

\item We also contribute a data set that satisfies the above two criteria from the domain of advanced steel property acquired by performing optical microscopy, comprising 4536 images. To date, no reported datasets relevant to heterogeneous tasks with subtle correlation for property modeling in MTL exist.
\end{itemize}


\section{Related work}
\label{sec:related literature}

\subsection{Multi-task learning networks}
\label{subsec:MultiTaskLearning}
For achieving performance with MTL networks, task-shared (generalizable) feature representations, and task-specific feature representations are both critical \cite{liu2019end}. The two requirements broadly translate to two key experimental knobs, namely network architecture and weighting scheme for loss function. 
Firstly, a network architecture (``how to share'') \cite{liu2019end} must permit task-shared (generalizable) features while providing the ability to learn task-specific features. Secondly, but equally critical, is the weighting of the loss function (``how to balance tasks?'') \cite{liu2019end}; it must enable learning all tasks without favoring only a few tasks towards learning completion while restricting others. The weight assignment to loss functions frequently demands automation \cite{liu2019end, chen2018gradnorm}; the tasks could comprise both classification and regression \cite{misra2016cross, liu2019end}, referred to as heterogeneous tasks.

\subsection{Architectures in multi-task learning}
\label{subsec:Architectures}
The network architecture evolution included several milestones. Hard Parameter Sharing (HPS) network architecture \cite{caruana1993multitask} involves sharing hidden layers across tasks while maintaining task-specific output layers. Cross-stitch network architecture \cite{misra2016cross} can model shared feature representation with linear combinations of feature maps that are end-to-end learnable via constituent cross-stitch units. Multi-Task Attention Network (MTAN) architecture \cite{liu2019end} automatically learns task-shared and task-specific features using a global feature pool with task-specific attention modules. For classification and regression, the networks perform multi-task learning for related tasks \cite{liu2019end, misra2016cross}. 

In contrast, for different levels of task-relatedness, Ma \etal proposed a Multi-gate Mixture of Experts (MMoE) architecture \cite{ma2018modeling} that can learn to model task relationships. Learning To Branch (LTB) architecture \cite{guo2020learning} learns ``where to share across tasks'' or ``where to perform task-specific branching'' in the network. Customized Gate Control (CGC) architecture \cite{tang2020progressive} involves a gating network to compute the weighted sum of task-shared and task-specific experts for extracting and separating deep semantic knowledge. It is unclear, however, if such architectures are effective for unrelated tasks bearing no correlation or only subtle correlation, including the added complexity of heterogeneous tasks involving classification and regression.  

\subsection{Weighting schemes for loss functions in multi-task learning}
\label{subsec:WeightingSchemes}

The weighting scheme is the second critical experimental knob, frequently adopted for MTL performance enhancement. Each scheme can consider a different feedback mechanism. Gradient-related feedback is one such widely used mechanism. Some examples include Gradient Normalization (GradNorm) \cite{chen2018gradnorm}, Multiple Gradient Descent Algorithm (MGDA) \cite{sener2018multi}, Gradient sign Dropout (GradDrop) \cite{chen2020just}, Projecting Conflicting Gradients (PCGrad) \cite{yu2020gradient}, Gradient Vaccine (GradVac) \cite{wang2020gradient}, Conflict Aversive Gradient (CAGard) \cite{liu2021conflict}, Multiple objective Convergence (MoCo) \cite{fernando2022mitigating}, Aligning of orthogonal components (Aligned MTL) \cite{senushkin2023independent}. 

Gradient feedback can consist of parameters such as positive curvature \cite{yu2020gradient}, average gradient \cite{yu2020gradient}, conflicting gradients \cite{yu2020gradient, chen2020just}, gradient differences \cite{yu2020gradient}, gradient magnitude similarity with cosine distance \cite{senushkin2023independent}, and gradient-based training rate adjustments for tasks \cite{chen2018gradnorm}. 

A recent report indicated the feasibility of scalarization, specifically ``smooth tchebycheff scalarization'' (STCH) for gradient-based multi-objective optimization \cite{lin2024smooth}. Recently, Impartial Multi-Task Learning (IMTL) \cite{liu2021towards} and Dual-Balancing for Multi-Task Learning (DB-MTL) \cite{lin2023dual} revealed the incorporation of not only gradient feedback but also loss feedback during weighting. 

Other feedback mechanisms for weighting include statistical Uncertainty Weighting (UW) \cite{kendall2018multi} uses homoscedastic uncertainty, Random Loss Weighting (RLW) \cite{lin2021reasonable}, Geometric Loss Strategy (GLS) \cite{chennupati2019multinet++} that is scale invariant. Some weights even consider the time aspect for weighting, like the Dynamic Weight Average (DWA) \cite{liu2019end}, or even consider game theory dynamics using Nash weighting (Nash MTL) \cite{navon2022multi}. Equal weighting (EW) \cite{lin2023libmtl} gets used for preliminary benchmarking. The choice of weighting scheme must ideally support learning on all tasks rather than its subset.

\subsection{Task grouping}
\label{subsec:LiteratureTaskGrouping}
For task grouping to avoid a negative transfer of knowledge between tasks, Standley \etal \cite{standley2020tasks} showed an empirical approach to determine relationships between jointly learned tasks. They attempted to find ``Which tasks should and should not be learned together in one network in MTL?''. Another task grouping approach, demonstrated by Fifty \etal \cite{fifty2021efficiently}, involved the quantification of inter-task affinity to systematically determine only those tasks for training together in an MTL network. Task grouping of heterogeneous tasks, with no more than a subtle correlation, can often be non-trivial in multi-task deep learning \cite{guo2020learning, tang2020progressive}, with only a few possibilities explored, primarily in architecture-weighting scheme landscape \cite{ott2022joint, zhang2014facial} that require substantial development to customize to use cases.

\subsection{Label transformation}
\label{subsec:LiteratureLabelTransformation}

Label transformation has been previously explored for tasks with correlation, even if heterogeneous tasks, comprising classification and regression. For example, in object detection and localization, the two tasks are related \cite{liu2019end, misra2016cross}, drawing in feature representations contributed majorly from a specific region of interest within an image. Optimal Transport Assignment (OTA) \cite{ge2021ota}, Task-aligned One-stage Object Detection (TOOD) \cite{feng2021tood}, and generalized Focal Loss (GFL) \cite{li2020generalized} are some examples that leveraged label transformation. Nevertheless, no literature on labeling strategies for dealing with unrelated or subtly correlated tasks involving classification and regression exists.

\subsection{Property modeling}
\label{subsec:LiteraturePropertyModeling}

Property prediction from images that occupy the entire image frame is a pervasive use case across science, engineering, and research. Some example disciplines for the use case include cell-biological sciences \cite{suresh2006mechanical}, petroleum science \cite{alqahtani2020machine}, medical science \cite{esteva2017dermatologist,ferreras2007diagnostic}, ecological science \cite{dobrowski2006improving}, materials science and engineering \cite{baux2020digitization, moravej2010electroformed}, mechanical engineering \cite{chen2024predicting}, remote sensing \cite{quiros2009testing}, applied mechanics \cite{li2019predicting}, and civil engineering \cite{mishra2019overview}. Although multi-task learning has recently been adopted for property modeling, these are primarily for correlated tasks \cite{iraki2024multi}. 

Given the breadth of impact of property modeling across disciplines, more generic approaches to handle task heterogeneity with no correlation or subtle correlation do not exist to date. There are also no image datasets in this context.

\begin{figure*}[tb]
  \centering
  \begin{subfigure}{0.48\linewidth}
  \centering
   \includegraphics[width=\textwidth]{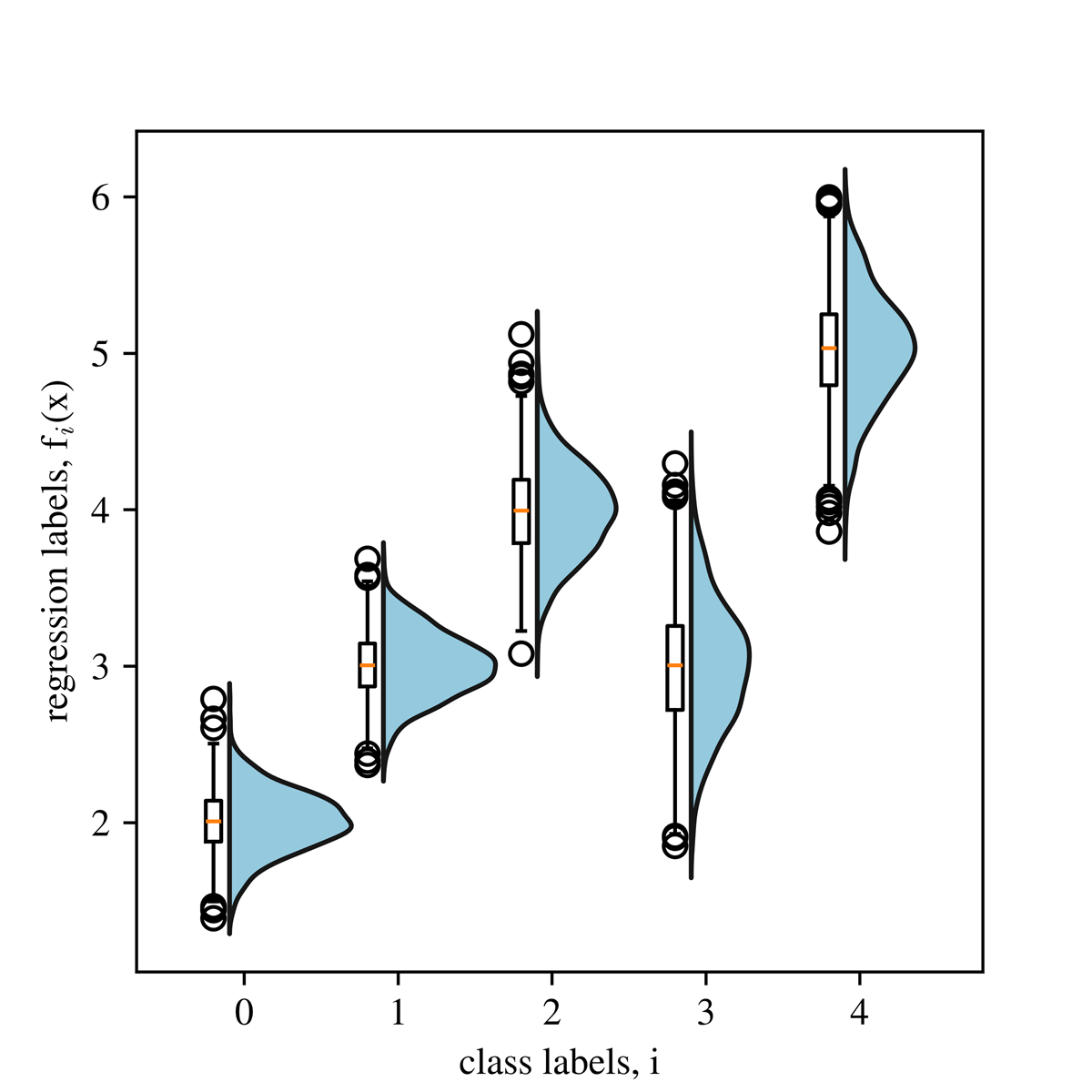}
    \caption{The regression labels, $f_i(x)$ can vary within a closed interval for an object class}
    \label{fig:short-a}
  \end{subfigure}
  \hfill
  \begin{subfigure}{0.48\linewidth}
  \centering
   \includegraphics[width=\textwidth]{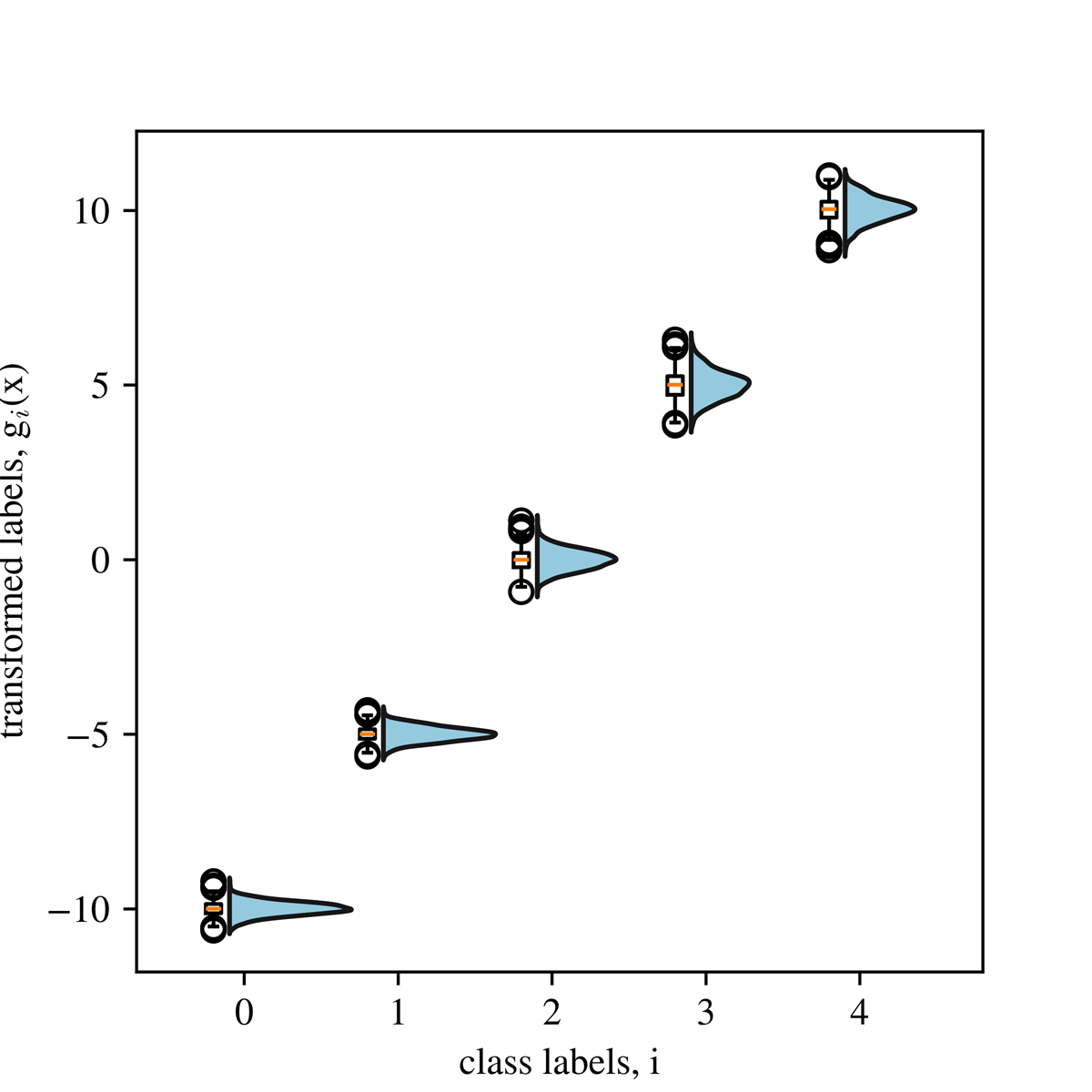}
    \caption{A transformed label, $g_i(x)$, contains the information about both the classification label and the regression label.}
    \label{fig:short-b}
  \end{subfigure}
  \caption{Schematic representation of regression labels and transformed labels (or hybrid labels) in a class-wise manner using a box plot and its distribution using a violin plot. Both regression labels and transformed labels belong to a continuous but distinct variable space. The transformed labels can allow learning heterogeneous tasks (a classification and a regression task) with only a subtle correlation particularly for property modeling via the FastCAR task consolidation approach.}
  \label{fig:short}
\end{figure*}

%% file: sec/2_formatting.tex



\section{FastCAR: Fast Classification And Regression}
\subsection{Property modeling}

Let $C = \{C_1, C_2, ..., C_n\}$ be the set of classes, where $n$ is the total number of classes.

For each class $C_i$, let $X_i$ be the set of instances of $C_i$.

Let $f_i: X_i \rightarrow \mathbb{R}$ be a property function that maps instances of $C_i$ to real numbers. 
For any instance $x \in X_i$:
\begin{equation}
    f_i(x) \in R_i \mid R_i = [a_i, b_i] = \{y \in \mathbb{R} : a_i \leq y \leq b_i\} 
\label{eq:property_function}    
\end{equation}

$f_i(x)$ is the property value, equivalent to regression label value, for an instance $x$ belonging to class $C_i$. The range of property values (range of regression labels) for instances of class $C_i$ can be expressed as a closed interval, where $a_i$ and $b_i$ are the lower and upper bounds of the closed interval, respectively.

The property model refers to a model that predicts the class label $C_i$ (or class index $i$) and the regression label $f_i(x)$ (the property value) of an instance $x$ of class $C_i$.

\subsection{FastCAR algorithm for label transformation}


We transform the label functions $f_i$ to $g_i$ such that the transformed labels are (a) disjoint (b) sorted (c) well separated according to \cref{eq:one-one-mapping,eq:monotonic_condition,eq:interclass_spacing}. Here, $g_i$ takes the form of $f_i + k_i$ for a well chosen $k_i$


 $g_i(x) \in S_i \mid S_i = R_i + k_i$ where $S_i$ is a closed interval just like in \cref{eq:property_function}
 

\begin{subequations}
 \begin{equation} 
 \begin{split}
  \forall p, q \in \{1, 2, ..., n\}, p \neq q:\\
  S_p \cap S_q = \emptyset
 \end{split}
 \label{eq:one-one-mapping}
 \end{equation}


\begin{algorithm}
\caption{FastCAR Label Transformation}
\begin{algorithmic}
\State \textbf{Input}: Set of class labels $C = \{C_1, C_2, ..., C_n\}$, set of regression labels or equivalent property functions $f_i: X_i \rightarrow \mathbb{R}$ .

\State \textbf{Output}: Transformed label $g_i(x)$

\State Initialize $n$ as the number of classes
\State Compute $[a_i, b_i]$ as the range of $f_i(x)$ for each class $C_i$
\State Set $\delta = \max_{i}\mid S_{i}\mid = \max_{i \in {1,2,...,n}}(b_i - a_i)$  
\State Set $u \in [1, 2] $, where $u$ is a hyperparameter 
\For{$i = 1$ to $n$}
\State $k_i \gets (i - 1) \cdot (u\cdot\delta)$, where $1 \leq u \leq 2$
\EndFor

\For{$i = 1$ to $n$}
\State $g_i(x) \gets f_i(x) + k_i$
\State $[a_i^g, b_i^g] \gets [a_i + k_i, b_i + k_i]$
\EndFor

\For{$i = 1$ to $n$}
\State $g_i(x) \gets g_i(x) - \frac{1}{2}\left(\max_{i} b_i^g - \min_{i} a_i^g\right)$ 
\EndFor

\State \Return $g_i(x)$

\end{algorithmic}
\end{algorithm}

For any two instances $x_p \in X_p$ and $x_q \in X_q$, where $p \neq q$ , i.e. instances from different classes, can not have the same property value in the transformed label space (hybrid label space) represented by \cref{eq:one-one-mapping}.
\cref{eq:monotonic_condition} prescribes the closed intervals of each class $S_i(x)$ to be sorted with reference to class index $i$, in the transformed label space (hybrid label space) 
 
 \begin{equation}
 \begin{split}
 S_1 < S_2 < ..... < S_{n-1} < S_n
 \end{split}
 \label{eq:monotonic_condition}
 \end{equation}

Note that the empty gap between consecutive ranges of transformed labels in \cref{fig:short-b}, i.e. between $S_{i+1}$ and $S_i$, lies between $\delta$ and $2\cdot\delta$, where $\delta$ represents  $\max_{i}\mid S_{i}\mid$ , i.e. the maximum range of any $S_i$. The condition has been represented using \cref{eq:interclass_spacing}. 

 \begin{equation}
  \begin{split}
  \forall i \in {1, 2, ..., n-1}: \\[-1ex]
  \delta < 
  \left(\min_{i}{S_{i+1}} - \max_{i}{S_i}\right) < 
  2\cdot\delta
 \end{split}
 \label{eq:interclass_spacing}
 \end{equation}
\end{subequations}

 \cref{eq:range_centering} shows range centering of the transformed labels.
 
\begin{equation}
g_i(x) = f_i(x) + k_i - \frac{1}{2}\left(\max_{i} b_i^g - \min_{i} a_i^g\right)
\label{eq:range_centering}
\end{equation}

\subsection{Rationale behind FastCAR algorithm design}

The formulation expressed in \cref{eq:one-one-mapping,eq:monotonic_condition}, represented visually in the schematic \cref{fig:short-b}, illustrates the possibility of unique mapping to a class index $i$, given the magnitude of transformed property $g(x)$. The formulation enables an efficient and compressed representation of both classification labels and regression labels via transformed labels. Previously, multi-task learning for heterogeneous tasks with subtle correlation necessitated both regression labels and classification labels for training and inference. 

\cref{eq:interclass_spacing} imposes a condition on the size of the empty spacing in the transformed label space between regression labels of successive class indices. Lower spacing increases the risk of misclassification, while significantly high spacing increases the risk of high gradients during backpropagation. Such gradient feedback gets implicitly incorporated while setting the coefficient of $\delta$ 

Range centering, shown in \cref{eq:range_centering}, can be particularly beneficial for negotiating high gradients during backpropagation. Such events are likely when the range of regression label values gets substantially wider or due to a significantly greater number of classes. Although not necessarily decisive for the present dataset, range-centering can generally benefit performance improvement.

\section{Experiments}
\label{sec:Experiments}


\textbf{Dataset.}
We created a dataset called ``Advanced Steel Property Dataset'' accessible via \url{https://github.com/fastcandr/Advanced-Steel-Property-Dataset}, relevant to the property modeling, highlighting the context of heterogeneous tasks with only a subtle correlation. The dataset consists of images representing instance of an object class occupies the entire image, annotated using a class label. Each instance of a detected object class is also associated with the property that can take continuous values, annotated using a regression label. The object classes indicate different steel material microstructures, while the continuous property variable represents hardness. 

The image acquisition scale and relevant scale for the measurement of object property must match; it was the microscopic length scale in this case, ensuring any further object localization is redundant. The dataset contained 4536 acquired images (resolution 224x224 pixels). A dataset size of about 4000 is typical in the domain of advanced steels \cite{lupulescu2015asm}. The dataset contained six object classes while maintaining class balance (an equal number of images for each class). Post-image acquisition, the property measurement using materials hardness testing was conducted, serving as regression labels. The dataset was split into training, validation, and testing sets in the ratio 5:1:1 while maintaining class balance.


\textbf{Evaluation metrics.} Performance evaluation of FastCAR on the classification task and the regression task, was conducted using classification accuracy and the mean squared error (MSE error). MTL models were used for benchmarking.  All models were granted 100 epochs, sufficient for the learning curve to flatten (validation loss curve), as confirmed in all cases. 

\begin{figure}[t]
  \centering
   \includegraphics[width=0.95\linewidth]{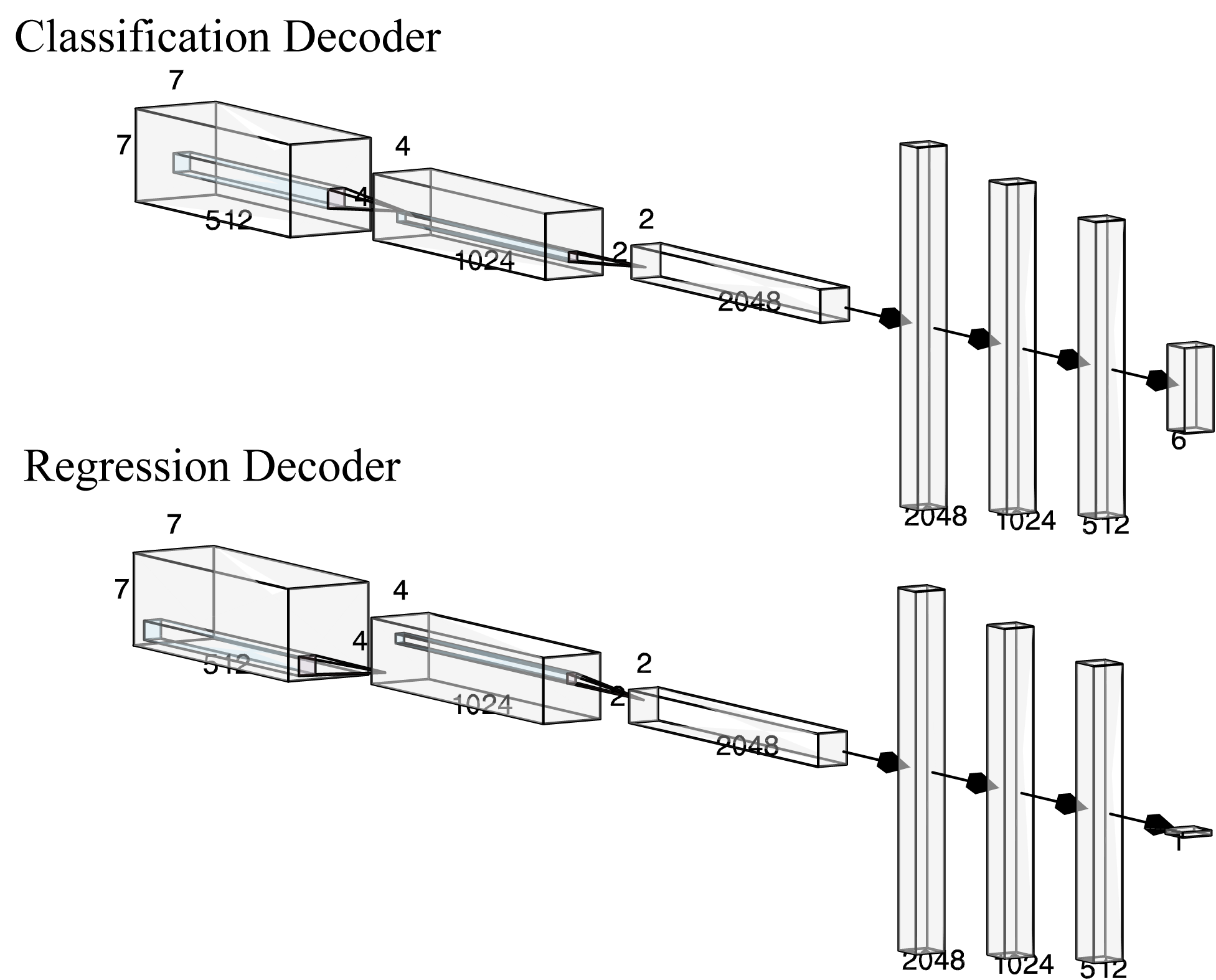}

   \caption{Decoder architecture of benchmark multi-task learning (MTL) models. The architecture comprises a classification decoder branch and a regression decoder branch, with shared feature representation using a Resnet-18 backbone as the input to the branches.}
   \label{fig:schematic_decoder}
\end{figure}

\textbf{Benchmark MTL models.} The MTL models for performance benchmarking consisted of 126 models per family, with two such families corresponding to 2 distinct decoder architectures. The first decoder consisted of architectural components between the feature extractor layer and the fully connected prediction layer (this final prediction layer is customizable for an individual task) of the Resnet18 network. 

The second decoder allows a more gradual data compression across successive layers of the network, unlike the previous decoder that allowed a very high data compression (98\%), with the possible risk of information loss (details in the supplementary section). Hence, the MTL model family associated with the second decoder was chosen in this study. \cref{fig:schematic_decoder} shows its architecture schematically. 

The MTL models contained combinations of 6 encoder architectures from \cref{subsec:Architectures}, 16 weighting schemes from \cref{subsec:WeightingSchemes}, and model hyperparameter variants. The pre-optimized hyperparameters or those that coincided with the default parameters prescribed in the LibMTL package \cite{lin2023libmtl} accounted for 84 models. The remaining 42 of the 126 models were further optimized by hyperparameter tuning. The best-performing MTL models from a multiplicative combination of 6 architectures and 16 weighting schemes, resulting in 96 MTL models, have been discussed subsequently.    

\textbf{Implementation details.} The FastCAR utilizes a ResNet-18-based regression network coupled with a label transformation algorithm. The benchmark MTL models comprised a Resnet-18 backbone. Maintaining a similar architecture between FastCAR and MTL models allowed us to isolate the effects of architectural dependence on performance, training time efficiency, and inference latency. The training of FastCAR and MTL families used an Adam optimizer with a learning rate of 1e-3 and a weight decay of 1e-4. A scheduler of type `reduced learning rate on the plateau' was used with a factor of 0.1, patience of 5, and verbose as True. All models were trained on an NVIDIA RTX\textsuperscript{TM} A6000 workstation with a dedicated GPU memory of 48 GB. A section on architectural parametrization has been visited in the supplementary section. 


\subsection{Performance of benchmark multi-task learning models}

\begin{figure*}[tb]
  \centering
  \includegraphics[height=7.65cm]{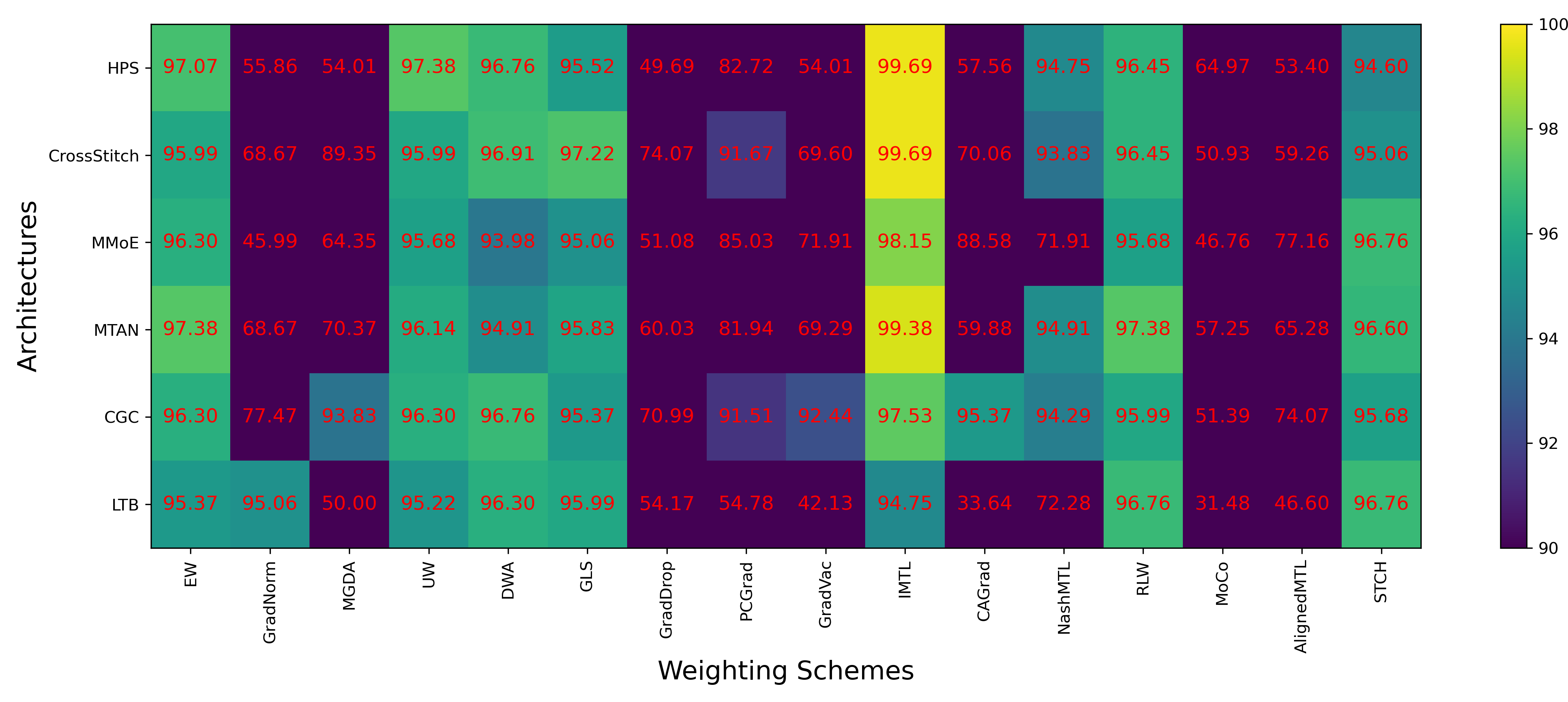}
  \caption{Performance of multi-task learning (MTL) models on the classification task (in terms of classification accuracy in \%) across the landscape of architecture \cite{misra2016cross,liu2019end, guo2020learning, ma2018modeling, caruana1993multitask, tang2020progressive} and weighting scheme \cite{liu2021conflict, liu2019end, yu2020gradient, wang2020gradient, chen2018gradnorm, senushkin2023independent, chen2020just, fernando2022mitigating, navon2022multi, sener2018multi, kendall2018multi, chennupati2019multinet++, lin2021reasonable, liu2021towards, lin2024smooth, lin2023libmtl}. The FastCAR achieves a classification accuracy of 99.54\%, which closely approaches the best model performance in the MTL landscape with 99.69\%.} 
  \label{fig:classification}
\end{figure*}

\begin{figure*}[tb]
  \centering
  \includegraphics[height=7.65cm]{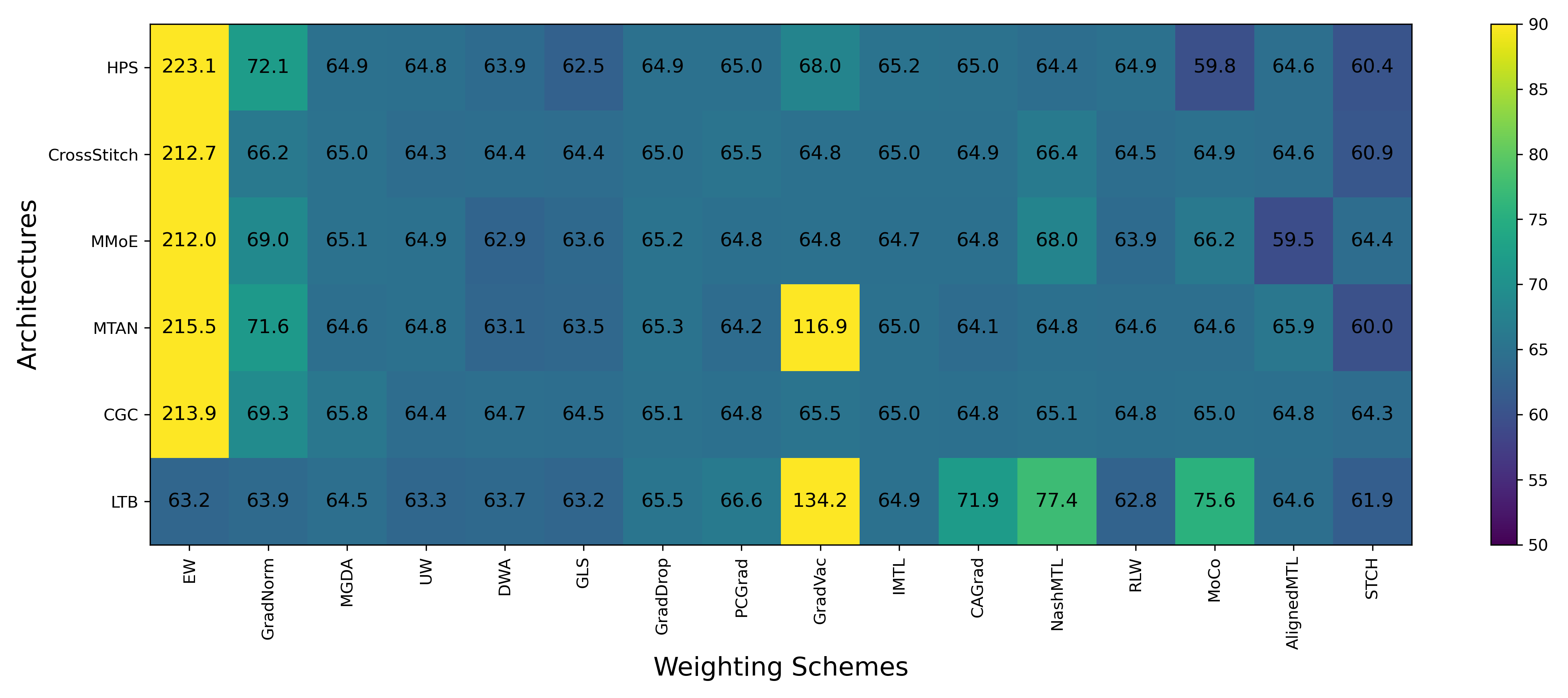}
  \caption{Performance of multi-task learning models on the regression task (in terms of MSE error expressed in multiples of 1000) across the landscape of architecture \cite{misra2016cross,liu2019end, guo2020learning, ma2018modeling, caruana1993multitask, tang2020progressive} and weighting scheme \cite{liu2021conflict, liu2019end, yu2020gradient, wang2020gradient, chen2018gradnorm, senushkin2023independent, chen2020just, fernando2022mitigating, navon2022multi, sener2018multi, kendall2018multi, chennupati2019multinet++, lin2021reasonable, liu2021towards, lin2024smooth, lin2023libmtl}. The error translates to a mean absolute percentage error of 54.2\% for the best performing MTL model, with most models approaching this value, while few remaining models resulted in even higher error. The FastCAR achieves a mean absolute percentage error of only 2.4\%.}
  \label{fig:regression}
\end{figure*}

The benchmark MTL models can learn the classification task in about half of the cases, as shown in \cref{fig:classification}. None of the MTL benchmark models could simultaneously learn the regression task \cref{fig:regression}. The best benchmark model, Cross stitch architecture with IMTL weighting, attained a classification accuracy of 99.69\%, but a high regression MSE error of 60 K (MAPE of over 55\%).    
The FastCAR achieves a classification accuracy of 99.54\% and a regression MSE error of about 0.5 K (MAPE of 2.4\%), confirming its feasibility for learning both the classification and the regression task.

\subsection{Why FastCAR performs better than multi-task learning model family?}
\label{WhyFastcarPerformsBetterthanMTL}

The use cases on object property modeling and object detection-localization share a commonality. The classification and regression tasks (heterogeneous tasks) can look at similar spatial regions of interest (ROI) in an image. The former holds the possibility that the ROI extends to the entire image frame. Due to this possible commonality, a shared feature representation cannot be directly ruled out. 

The MTL models (a total of 252 models corresponding to two families) do not succeed in learning both classification and regression tasks despite parameterizing in the landscape of architecture and weighting scheme. From an architectural viewpoint, even complex architectures with attention modules like MTAN do not play a supporting role in offering shared representations that are acceptable for weighting schemes. The weighing schemes also do not offer enough to the relationship with the architectures to harmoniously learn both tasks. 

Based on the large MSE error magnitude observed for the regression task across MTL models, a mismatch of the relative scale of the weights between classification and regression tasks is possible. The Geometric Loss Scheme (GLS) has the potential to remain scale-invariant, involving both classification and regression, which also does not appear to serve presently for property modeling. 

The property modeling of object classes is associated with subtle task correlations between heterogeneous tasks. The two tasks need not necessarily honor the possibility of sharing the entire image-ROI for feature representation \cite{standley2020tasks}. The feature space of individual tasks need not be similar and can lack a shared representation. 

In this context, architectures involving varying extents of task correlation, like LTB and CGC, can learn task relationships to a certain extent. The MTL model performance indicates a plausible mismatch in the learning ability between the tasks, which is beyond the comprehension of architecture and the weighting scheme landscape for any meaningful shared features. Architectural complexity, for instance, attention mechanism-based architectures, do not necessarily imply the potential to solve the complexity of task heterogeneity with subtle correlation. These are two different dimensions of complexity for permitting a direct comparison. Note that the MTAN architectures with attention modules reported are for heterogeneous and related tasks. 

The FastCAR design can demonstrate performance on classification and regression tasks (heterogeneous tasks) despite the non-trivial challenge of subtle task correlation. The current approach has been equipped to handle the classification task by minimizing misclassifications and also implicitly handling gradient feedback associated with regression. Neither explicit weight assignment for the loss functions for individual heterogeneous tasks nor a robust encoder architecture is needed. 

\subsection{Ablation study}

An ablative study, as shown in \cref{table:ablation}, indicates the experimental variables and their effects on the performance. A combination of neural network architecture and transformed labels is critical; more specifically, a prediction layer with one neuron (a regression network) and ``Good'' labels achieve learning on both heterogeneous tasks, involving the classification task and the regression task, despite only a subtle correlation between them. Note that transformed labels (or hybrid labels) are defined as `Good' labels if \cref{eq:one-one-mapping,eq:monotonic_condition,eq:interclass_spacing} are obeyed, and `Bad' labels otherwise. Range-centering can be generically beneficial to negotiate high gradients if encountered in the network during backpropagation, despite not being decisive presently.

\begin{table*}[]
\centering

\begin{tabular}{@{}ccc|cc@{}}
\toprule
\multicolumn{3}{c|}{Experimental design variables}                                                                                                                                                                                                                                        & \multicolumn{2}{c}{Performance}                                                                                                                                                         \\ \midrule
\multicolumn{1}{c|}{\multirow{2}{*}{\begin{tabular}[c]{@{}c@{}}neurons \\ in prediction layer of \\ single-task network\end{tabular}}} & \multicolumn{2}{c|}{transformed labels}                                                                                                               & \multirow{2}{*}{\begin{tabular}[c]{@{}c@{}}classification\\ accuracy \\ (\%) \\ $\uparrow$ \end{tabular}} & \multirow{2}{*}{\begin{tabular}[c]{@{}c@{}}regression \\ error\\ (MSE, MAPE)\\ $\downarrow$ \end{tabular}} \\ \cmidrule(lr){2-3}
\multicolumn{1}{c|}{}                                                                                                                  & \begin{tabular}[c]{@{}c@{}}$k_i$\\ \\ (Good / Bad)\end{tabular} & \begin{tabular}[c]{@{}c@{}}g(x) \\ range centering\\ (True/False)\end{tabular} &                                                                                            &                                                                                            \\ \midrule
\multicolumn{1}{c|}{1}                                                                                                                 & Good                                                            & True                                                                           & 99.54\%                                                                                    & 0.52 K, 2.6\%                                                                              \\
\multicolumn{1}{c|}{1}                                                                                                                 & Good                                                            & False                                                                          & 99.54\%                                                                                    & 0.43 K, 2.4\%                                                                              \\
\multicolumn{1}{c|}{1}                                                                                                                 & Bad                                                             & True                                                                           & 18.82\%                                                                                    & 0.15 K, 1.9\%                                                                              \\
\multicolumn{1}{c|}{1}                                                                                                                 & Bad                                                             & False                                                                          & 18.67\%                                                                                    & 0.18 K, 2.0\%                                                                              \\ \midrule
\multicolumn{1}{c|}{\textgreater{}1}                                                                                                   & Good                                                            & True                                                                           & \multicolumn{2}{c}{\multirow{4}{*}{N/A}}                                                                                                                                                \\
\multicolumn{1}{c|}{\textgreater{}1}                                                                                                   & Good                                                            & False                                                                          & \multicolumn{2}{c}{}                                                                                                                                                                    \\
\multicolumn{1}{c|}{\textgreater{}1}                                                                                                   & Bad                                                             & True                                                                           & \multicolumn{2}{c}{}                                                                                                                                                                    \\
\multicolumn{1}{c|}{\textgreater{}1}                                                                                                   & Bad                                                             & False                                                                          & \multicolumn{2}{c}{}                                                                                                                                                                    \\ \midrule
\multicolumn{1}{c|}{\begin{tabular}[c]{@{}c@{}}Best MTL benchmark \\ (Cross-stitch architecture \\ with IMTL weighting)\end{tabular}} & \multicolumn{2}{c|}{N/A}                                                                                                                         & 99.69\%                                                                                    & \multicolumn{1}{l}{65 K,  57.5\%}                                                            \\ \bottomrule
\end{tabular}
\caption{An ablation study revealed the effect of experimental design variables on the performance metrics for different combinations of network configuration and transformed labels. A deep learning network with one neuron in the prediction layer (regression network) in combination with appropriately transformed labels, complying with \cref{eq:one-one-mapping,eq:monotonic_condition,eq:interclass_spacing} (``Good'' labels) and \cref{eq:range_centering}, achieve learning with FastCAR on both heterogeneous tasks despite only subtly correlated. In the ``Performance'' column, $\uparrow$ indicates higher is better, while $\downarrow$ implies the lower is better.}
\label{table:ablation}
\end{table*}

\subsection{How fast is FastCAR? and why is it fast?}
\label{Time}

\cref{table:TrainingInferenceTime} shows the training time duration (in minutes) and the inference time (in seconds) for various benchmark MTL architectures parametrized across weighting schemes. FastCAR requires substantially lower training time (2.52x quicker) and lower inference latency (55\% faster) when compared with the best-performing MTL model, corresponding to Cross-stitch architecture with an IMTL weighting scheme. 

Given that MTL models are often associated with reduced latency \cite{guo2020learning, sagduyu2024low}, the inference latency improvement of 55\% with FastCAR is significant compared to the best-performing MTL model. The latency improvement could be as high as about 135\% over CGC architectures. A similar trend was observed with FastCAR for training time efficiency approaching 7.5x quicker than an MTL model with CGC architecture and GradVac weighting scheme combination.      

The label transformation currently takes the form of a simple linear transformation. This formulation solves both tasks with efforts equivalent to one task via a single-task regression network, unlike MTL networks frequently requiring bulky encoder architectures.



\section{Conclusions}

FastCAR serves as a fundamental building block for task consolidation of a classification task and a regression task (heterogeneous tasks) despite the non-triviality of subtle task correlation via label transformation for property modeling in multi-task learning (MTL). The use case involves the classification of a detected object instance (occupying the entire image frame) and regression modeling of its property that can take continuous values, a pervasive use case across scientific and engineering disciplines.

The transformed label formulation permits learning of the classification task by minimizing misclassification (classification accuracy of 99.45\%). It also learns the regression task effectively (mean absolute percentage error of 2.4\%) by incorporating implicit gradient feedback. The learning on both tasks collectively was a challenge for benchmark MTL models parametrized in the landscape of established architectures and weighting schemes (252 MTL models).  

The label transformation can effectively represent classification labels and regression labels in a compressed and transformed representation. The unique identification of the class index from the transformed label is the key principle behind this formulation.  

FastCAR is time efficient due to its computationally inexpensive labeling strategy and its feasibility for combining with a deep learning regression network. It allows training time efficiency (2.52x quicker) and reduced latency (55\% faster) compared to the best-performing benchmark MTL model. The label transformation involves a simple linear transformation to solve both tasks, with efforts equivalent to one task via a single-task regression network, unlike MTL networks necessitating bulky architectures.  

A key limitation of this work, while providing scope for future research, is that a greater generic utility of FastCAR, beyond those for broad and diverse property modeling use cases, requires a problem statement to be necessarily formulated in terms of object-property. 

%% file: sec/X_suppl.tex
\clearpage
\setcounter{page}{1}
\maketitlesupplementary

\section{Supplementary section}
\subsection{Empirical guideline for FastCAR algorithm}
\label{sec:rationale}

Following empirically determined guideline was used for assessing the  progress of FastCAR algorithm for learning both tasks, in terms of validation loss and average gradient across trainable parameters of the network. 

\begin{subequations}
\begin{equation}
\min_{1 < i \leq 20} \frac{\text{Validation Loss}_{\text{epoch} = i}}{\text{Validation Loss}_{\text{epoch} = 1}} \leq \frac{1}{2}
\label{eq:loss}
\end{equation}
\begin{equation}
\max_{1 < i \leq 20} \frac{\text{Average Gradient}_{\text{epoch} = i}}{\text{Average Gradient}_{\text{epoch} = 1}} \geq 2
\label{eq:gradient}
\end{equation}
\end{subequations}

\subsection{Hyperparameters for benchmark multi-task learning models}
The model hyperparameters for weighting schemes were performed for 42 benchmark MTL models out of the 126 models per family. The parameterization was prominent for MGDA and STCH weighting schemes. Note that the MGDA scheme included three variants, `loss', `loss +', and `L2', while the STCH included four variants varied over $\mu$ of 0.01, 0.1, 0.5, and 1.

\subsection{Decoder architectures variants corresponding to each MTL model family}
Two MTL model families for benchmarking correspond to the following two decoder configurations. The Decoder-1 contains one more fully connected (fc) layer with 512 neurons (along with 1D batch normalization, ReLU activation function, and dropout of 0.5), compared to a ResNet-18 pre-trained network without the backbone. The additional layer lies between the already present adaptive average pooling layer and the fully connected layer. 

The Decoder-2, compared to Decoder-1, contains two more convolutional blocks (each followed by batch normalization and ReLU) located before the adaptive average pooling layer. The convolutional blocks have feature maps of sizes 4×4 followed by 2×2, with channels doubling across each convolutional block. After the pooling layer, yet another fully connected layer was placed.   

The experimental design focussed on serving a dual purpose. First, benchmarking of the FastCAR model performance. Second, to search and identify MTL models that can perform despite possible heterogeneous behavior between tasks (which are often challenging to determine explicitly), by attempting various combinations of architectures and weighting schemes without an explicit assignment. 

\subsection{Choice of decoder architecture and data compression}
From a data compression perspective, across consecutive layers (convolutional blocks) of the ResNet-18 network, the compression factor does not exceed 2 (50\% compression); the feature maps size decreases by two times along height and width, despite an increase in channels by two times. However, after layer 4 (as designated in ResNet-18), there is substantial data compression by 49 times (~98\% compression) during adaptive average 2D pooling. The feature map size decreases to 1/7\textsuperscript{th} of height and width (feature map size decreases to 1/49\textsuperscript{th}) with an unaltered number of channels before entering the adaptive average pooling. Adaptive average pooling often eliminates topological constraints due to mismatching dimensions of successive convolutional blocks \cite{van2019evolutionary}. It is unclear if such pooling, particularly in Decoder-1, can retain important information from the shared feature representation despite substantial compression, particularly in the context of MTL.

The Decoder2 configuration, as compared to the ResNet-18 backbone architecture, restricts the data compression factor to less than 2, even after layer 4. In other words, more than 50\% compression per layer gets prevented during the adaptive average pooling. In addition to the increased potential to tackle information loss during compression, Decoder-2 provides additional learnable parameters.

\subsection{Parametrization of backbone architectures for MTL and FastCAR} 

More significantly, the parametrization of architectures in MTL models did not accomplish learning both tasks while also faced with exceeding the memory limits of the device; the models were trained on an NVIDIA RTX\textsuperscript{TM} A6000 workstation with a dedicated GPU memory of 48 GB. Hence, a direct comparison with FastCAR is not feasible. The backbone architectural parametrization of FastCAR approached saturation even with the simplest of Resnet architecture or the Resnet18 architecture. Other variants like Resnet32,  Resnet50, and Resnet100 were also attempted with marginal benefits at the expense of time efficiency. 

\subsection{Training duration of MTL models}
The type of MTL architecture and the type of weighting scheme collectively influence the training time. Yet, the architecture plays a decisive role in the current problem statement. The decisiveness of MTL architectures over the weighting schemes was even more prominent for latency evaluation via inference time. A similar trend is noticeable with all MTL architectures towards the inference time, with lower variance (or lower ratio of standard deviation to mean) than that for training time. 

FastCAR portrays feasibility demonstration involving performance, achieves training time efficiency, and allows reduced latency compared to those with the MTL model family. 

\begin{table}[ht]
    \centering
    \small
    \begin{tabular}{lcc}
        \toprule
        \multirow{2}{*}{\textbf{MTL Architecture}} & \textbf{Train Time (min)} & \textbf{Inference Time (sec)} \\
        & & \\
        \midrule
        HPS & $24.638 \pm 7.440$ & $1.289 \pm 0.028$ \\
        Cross-stitch & $39.749 \pm 14.203$ & $1.936 \pm 0.023$ \\
        MMoE & $39.511 \pm 13.889$ & $1.927 \pm 0.008$ \\
        MTAN & $36.338 \pm 13.641$ & $1.571 \pm 0.007$ \\
        CGC & $78.037 \pm 35.674$ & $2.921 \pm 0.013$ \\
        LTB & $38.470 \pm 12.735$ & $1.960 \pm 0.011$ \\
        \midrule
        FastCAR & $16.320 \pm 0.142$ & $1.241 \pm 0.005$ \\
        \bottomrule
    \end{tabular}
    \caption{Training time duration (min) and inference time duration (sec) of benchmark multi-task network architectures (parametrized over various weighting schemes). FastCAR suffices a substantially lower training time (2.52x quicker) and lower inference latency (55\% faster) than the best observed MTL model in terms of performance (Cross-stitch architecture with IMTL weighting).}
    \label{table:TrainingInferenceTime}
\end{table}

%% file: main.bbl
\begin{thebibliography}{48}
\providecommand{\natexlab}[1]{#1}
\providecommand{\url}[1]{\texttt{#1}}
\expandafter\ifx\csname urlstyle\endcsname\relax
  \providecommand{\doi}[1]{doi: #1}\else
  \providecommand{\doi}{doi: \begingroup \urlstyle{rm}\Url}\fi

\bibitem[Alqahtani et~al.(2020)Alqahtani, Alzubaidi, Armstrong, Swietojanski, and Mostaghimi]{alqahtani2020machine}
Naif Alqahtani, Fatimah Alzubaidi, Ryan~T Armstrong, Pawel Swietojanski, and Peyman Mostaghimi.
\newblock Machine learning for predicting properties of porous media from 2d x-ray images.
\newblock \emph{Journal of Petroleum Science and Engineering}, 184:\penalty0 106514, 2020.

\bibitem[Baux et~al.(2020)Baux, Cou{\'e}gnat, Vignoles, Lasseux, Kuhn, Carucci, Mano, and Le]{baux2020digitization}
Anthony Baux, Guillaume Cou{\'e}gnat, G{\'e}rard~L Vignoles, Didier Lasseux, Alexander Kuhn, Cristina Carucci, Nicolas Mano, and Tien~Dung Le.
\newblock Digitization and image-based structure-properties relationship evaluation of a porous gold micro-electrode.
\newblock \emph{Materials \& Design}, 193:\penalty0 108812, 2020.

\bibitem[Caruana(1993)]{caruana1993multitask}
R Caruana.
\newblock Multitask learning: A knowledge-based source of inductive bias1.
\newblock In \emph{Proceedings of the Tenth International Conference on Machine Learning}, pages 41--48. Citeseer, 1993.

\bibitem[Chen et~al.(2024)Chen, Tang, Liu, Liu, and Zhou]{chen2024predicting}
Haolong Chen, Xinyue Tang, Zhaotao Liu, Zhanli Liu, and Huanlin Zhou.
\newblock Predicting the temperature field of thermal cloaks in homogeneous isotropic multilayer materials based on deep learning.
\newblock \emph{International Journal of Heat and Mass Transfer}, 219:\penalty0 124849, 2024.

\bibitem[Chen et~al.(2018)Chen, Badrinarayanan, Lee, and Rabinovich]{chen2018gradnorm}
Zhao Chen, Vijay Badrinarayanan, Chen-Yu Lee, and Andrew Rabinovich.
\newblock Gradnorm: Gradient normalization for adaptive loss balancing in deep multitask networks.
\newblock In \emph{International conference on machine learning}, pages 794--803. PMLR, 2018.

\bibitem[Chen et~al.(2020)Chen, Ngiam, Huang, Luong, Kretzschmar, Chai, and Anguelov]{chen2020just}
Zhao Chen, Jiquan Ngiam, Yanping Huang, Thang Luong, Henrik Kretzschmar, Yuning Chai, and Dragomir Anguelov.
\newblock Just pick a sign: Optimizing deep multitask models with gradient sign dropout.
\newblock \emph{Advances in Neural Information Processing Systems}, 33:\penalty0 2039--2050, 2020.

\bibitem[Chennupati et~al.(2019)Chennupati, Sistu, Yogamani, and A~Rawashdeh]{chennupati2019multinet++}
Sumanth Chennupati, Ganesh Sistu, Senthil Yogamani, and Samir A~Rawashdeh.
\newblock Multinet++: Multi-stream feature aggregation and geometric loss strategy for multi-task learning.
\newblock In \emph{Proceedings of the IEEE/CVF Conference on Computer Vision and Pattern Recognition Workshops}, pages 0--0, 2019.

\bibitem[Di~Fatta et~al.(2023)Di~Fatta, Nicosia, Ojha, and Pardalos]{di2023multi}
Giuseppe Di~Fatta, Giuseppe Nicosia, Varun Ojha, and Panos Pardalos.
\newblock Multi-task deep learning as multi-objective optimization.
\newblock In \emph{Encyclopedia of Optimization}, pages 1--10. Springer, 2023.

\bibitem[Dobrowski et~al.(2006)Dobrowski, Greenberg, Ramirez, and Ustin]{dobrowski2006improving}
SZ Dobrowski, JA Greenberg, CM Ramirez, and SL Ustin.
\newblock Improving image derived vegetation maps with regression based distribution modeling.
\newblock \emph{Ecological Modelling}, 192\penalty0 (1-2):\penalty0 126--142, 2006.

\bibitem[Esteva et~al.(2017)Esteva, Kuprel, Novoa, Ko, Swetter, Blau, and Thrun]{esteva2017dermatologist}
Andre Esteva, Brett Kuprel, Roberto~A Novoa, Justin Ko, Susan~M Swetter, Helen~M Blau, and Sebastian Thrun.
\newblock Dermatologist-level classification of skin cancer with deep neural networks.
\newblock \emph{nature}, 542\penalty0 (7639):\penalty0 115--118, 2017.

\bibitem[Feng et~al.(2021)Feng, Zhong, Gao, Scott, and Huang]{feng2021tood}
Chengjian Feng, Yujie Zhong, Yu Gao, Matthew~R Scott, and Weilin Huang.
\newblock Tood: Task-aligned one-stage object detection.
\newblock In \emph{2021 IEEE/CVF International Conference on Computer Vision (ICCV)}, pages 3490--3499. IEEE Computer Society, 2021.

\bibitem[Fernando et~al.(2022)Fernando, Shen, Liu, Chaudhury, Murugesan, and Chen]{fernando2022mitigating}
Heshan~Devaka Fernando, Han Shen, Miao Liu, Subhajit Chaudhury, Keerthiram Murugesan, and Tianyi Chen.
\newblock Mitigating gradient bias in multi-objective learning: A provably convergent approach.
\newblock In \emph{The Eleventh International Conference on Learning Representations}, 2022.

\bibitem[Ferreras et~al.(2007)Ferreras, Pajar{\'\i}n, Polo, Larrosa, Pablo, and Honrubia]{ferreras2007diagnostic}
Antonio Ferreras, Ana~B Pajar{\'\i}n, Vicente Polo, Jos{\'e}~M Larrosa, Lu{\'\i}s~E Pablo, and Francisco~M Honrubia.
\newblock Diagnostic ability of heidelberg retina tomograph 3 classifications: glaucoma probability score versus moorfields regression analysis.
\newblock \emph{Ophthalmology}, 114\penalty0 (11):\penalty0 1981--1987, 2007.

\bibitem[Fifty et~al.(2021)Fifty, Amid, Zhao, Yu, Anil, and Finn]{fifty2021efficiently}
Chris Fifty, Ehsan Amid, Zhe Zhao, Tianhe Yu, Rohan Anil, and Chelsea Finn.
\newblock Efficiently identifying task groupings for multi-task learning.
\newblock \emph{Advances in Neural Information Processing Systems}, 34:\penalty0 27503--27516, 2021.

\bibitem[Ge et~al.(2021)Ge, Liu, Li, Yoshie, and Sun]{ge2021ota}
Zheng Ge, Songtao Liu, Zeming Li, Osamu Yoshie, and Jian Sun.
\newblock Ota: Optimal transport assignment for object detection.
\newblock In \emph{Proceedings of the IEEE/CVF Conference on Computer Vision and Pattern Recognition}, pages 303--312, 2021.

\bibitem[Guo et~al.(2020)Guo, Lee, and Ulbricht]{guo2020learning}
Pengsheng Guo, Chen-Yu Lee, and Daniel Ulbricht.
\newblock Learning to branch for multi-task learning.
\newblock In \emph{International conference on machine learning}, pages 3854--3863. PMLR, 2020.

\bibitem[Iraki et~al.(2024)Iraki, Morand, Dornheim, Link, and Helm]{iraki2024multi}
Tarek Iraki, Lukas Morand, Johannes Dornheim, Norbert Link, and Dirk Helm.
\newblock A multi-task learning-based optimization approach for finding diverse sets of microstructures with desired properties.
\newblock \emph{Journal of Intelligent Manufacturing}, 35\penalty0 (4):\penalty0 1887--1903, 2024.

\bibitem[Kendall et~al.(2018)Kendall, Gal, and Cipolla]{kendall2018multi}
Alex Kendall, Yarin Gal, and Roberto Cipolla.
\newblock Multi-task learning using uncertainty to weigh losses for scene geometry and semantics.
\newblock In \emph{Proceedings of the IEEE conference on computer vision and pattern recognition}, pages 7482--7491, 2018.

\bibitem[Li et~al.(2019)Li, Liu, Cui, Luo, Li, and Zhuang]{li2019predicting}
Xiang Li, Zhanli Liu, Shaoqing Cui, Chengcheng Luo, Chenfeng Li, and Zhuo Zhuang.
\newblock Predicting the effective mechanical property of heterogeneous materials by image based modeling and deep learning.
\newblock \emph{Computer Methods in Applied Mechanics and Engineering}, 347:\penalty0 735--753, 2019.

\bibitem[Li et~al.(2020)Li, Wang, Wu, Chen, Hu, Li, Tang, and Yang]{li2020generalized}
Xiang Li, Wenhai Wang, Lijun Wu, Shuo Chen, Xiaolin Hu, Jun Li, Jinhui Tang, and Jian Yang.
\newblock Generalized focal loss: Learning qualified and distributed bounding boxes for dense object detection.
\newblock \emph{Advances in Neural Information Processing Systems}, 33:\penalty0 21002--21012, 2020.

\bibitem[Lin and Zhang(2023)]{lin2023libmtl}
Baijiong Lin and Yu Zhang.
\newblock Libmtl: A python library for deep multi-task learning.
\newblock \emph{Journal of Machine Learning Research}, 24\penalty0 (1-7):\penalty0 18, 2023.

\bibitem[Lin et~al.(2021)Lin, Ye, Zhang, and Tsang]{lin2021reasonable}
Baijiong Lin, Feiyang Ye, Yu Zhang, and Ivor~W Tsang.
\newblock Reasonable effectiveness of random weighting: A litmus test for multi-task learning.
\newblock \emph{arXiv preprint arXiv:2111.10603}, 2021.

\bibitem[Lin et~al.(2023)Lin, Jiang, Ye, Zhang, Chen, Chen, Liu, and Kwok]{lin2023dual}
Baijiong Lin, Weisen Jiang, Feiyang Ye, Yu Zhang, Pengguang Chen, Ying-Cong Chen, Shu Liu, and James~T Kwok.
\newblock Dual-balancing for multi-task learning.
\newblock \emph{arXiv preprint arXiv:2308.12029}, 2023.

\bibitem[Lin et~al.(2024)Lin, Zhang, Yang, Liu, Wang, and Zhang]{lin2024smooth}
Xi Lin, Xiaoyuan Zhang, Zhiyuan Yang, Fei Liu, Zhenkun Wang, and Qingfu Zhang.
\newblock Smooth tchebycheff scalarization for multi-objective optimization.
\newblock \emph{arXiv preprint arXiv:2402.19078}, 2024.

\bibitem[Liu et~al.(2021{\natexlab{a}})Liu, Liu, Jin, Stone, and Liu]{liu2021conflict}
Bo Liu, Xingchao Liu, Xiaojie Jin, Peter Stone, and Qiang Liu.
\newblock Conflict-averse gradient descent for multi-task learning.
\newblock \emph{Advances in Neural Information Processing Systems}, 34:\penalty0 18878--18890, 2021{\natexlab{a}}.

\bibitem[Liu et~al.(2021{\natexlab{b}})Liu, Li, Kuang, Xue, Chen, Yang, Liao, and Zhang]{liu2021towards}
Liyang Liu, Yi Li, Zhanghui Kuang, J Xue, Yimin Chen, Wenming Yang, Qingmin Liao, and Wayne Zhang.
\newblock Towards impartial multi-task learning.
\newblock In \emph{Towards impartial multi-task learning}. iclr, 2021{\natexlab{b}}.

\bibitem[Liu et~al.(2019)Liu, Johns, and Davison]{liu2019end}
Shikun Liu, Edward Johns, and Andrew~J Davison.
\newblock End-to-end multi-task learning with attention.
\newblock In \emph{Proceedings of the IEEE/CVF conference on computer vision and pattern recognition}, pages 1871--1880, 2019.

\bibitem[Lupulescu et~al.(2015)Lupulescu, Flowers, Vermillion, and Henry]{lupulescu2015asm}
Afina Lupulescu, Tyler Flowers, Linda Vermillion, and Scott Henry.
\newblock Asm micrograph database™.
\newblock \emph{Metallography, Microstructure, and Analysis}, 4:\penalty0 322--327, 2015.

\bibitem[Ma et~al.(2018)Ma, Zhao, Yi, Chen, Hong, and Chi]{ma2018modeling}
Jiaqi Ma, Zhe Zhao, Xinyang Yi, Jilin Chen, Lichan Hong, and Ed~H Chi.
\newblock Modeling task relationships in multi-task learning with multi-gate mixture-of-experts.
\newblock In \emph{Proceedings of the 24th ACM SIGKDD international conference on knowledge discovery \& data mining}, pages 1930--1939, 2018.

\bibitem[Mishra and Singh(2019)]{mishra2019overview}
Onkar Mishra and SP Singh.
\newblock An overview of microstructural and material properties of ultra-high-performance concrete.
\newblock \emph{Journal of Sustainable Cement-Based Materials}, 8\penalty0 (2):\penalty0 97--143, 2019.

\bibitem[Misra et~al.(2016)Misra, Shrivastava, Gupta, and Hebert]{misra2016cross}
Ishan Misra, Abhinav Shrivastava, Abhinav Gupta, and Martial Hebert.
\newblock Cross-stitch networks for multi-task learning.
\newblock In \emph{Proceedings of the IEEE conference on computer vision and pattern recognition}, pages 3994--4003, 2016.

\bibitem[Moravej et~al.(2010)Moravej, Prima, Fiset, and Mantovani]{moravej2010electroformed}
M Moravej, F Prima, M Fiset, and D Mantovani.
\newblock Electroformed iron as new biomaterial for degradable stents: Development process and structure--properties relationship.
\newblock \emph{Acta biomaterialia}, 6\penalty0 (5):\penalty0 1726--1735, 2010.

\bibitem[Navon et~al.(2022)Navon, Shamsian, Achituve, Maron, Kawaguchi, Chechik, and Fetaya]{navon2022multi}
Aviv Navon, Aviv Shamsian, Idan Achituve, Haggai Maron, Kenji Kawaguchi, Gal Chechik, and Ethan Fetaya.
\newblock Multi-task learning as a bargaining game.
\newblock \emph{arXiv preprint arXiv:2202.01017}, 2022.

\bibitem[Ott et~al.(2022)Ott, R{\"u}gamer, Heublein, Bischl, and Mutschler]{ott2022joint}
Felix Ott, David R{\"u}gamer, Lucas Heublein, Bernd Bischl, and Christopher Mutschler.
\newblock Joint classification and trajectory regression of online handwriting using a multi-task learning approach.
\newblock In \emph{Proceedings of the IEEE/CVF winter conference on applications of computer vision}, pages 266--276, 2022.

\bibitem[Quir{\'o}s et~al.(2009)Quir{\'o}s, Felic{\'\i}simo, and Cuartero]{quiros2009testing}
Elia Quir{\'o}s, {\'A}ngel~M Felic{\'\i}simo, and Aurora Cuartero.
\newblock Testing multivariate adaptive regression splines (mars) as a method of land cover classification of terra-aster satellite images.
\newblock \emph{Sensors}, 9\penalty0 (11):\penalty0 9011--9028, 2009.

\bibitem[Sagduyu et~al.(2024)Sagduyu, Erpek, Yener, and Ulukus]{sagduyu2024low}
Yalin~E Sagduyu, Tugba Erpek, Aylin Yener, and Sennur Ulukus.
\newblock Low-latency task-oriented communications with multi-round, multi-task deep learning.
\newblock In \emph{Proceedings of the 30th Annual International Conference on Mobile Computing and Networking}, pages 2365--2370, 2024.

\bibitem[Sener and Koltun(2018)]{sener2018multi}
Ozan Sener and Vladlen Koltun.
\newblock Multi-task learning as multi-objective optimization.
\newblock \emph{Advances in neural information processing systems}, 31, 2018.

\bibitem[Senushkin et~al.(2023)Senushkin, Patakin, Kuznetsov, and Konushin]{senushkin2023independent}
Dmitry Senushkin, Nikolay Patakin, Arseny Kuznetsov, and Anton Konushin.
\newblock Independent component alignment for multi-task learning.
\newblock In \emph{Proceedings of the IEEE/CVF Conference on Computer Vision and Pattern Recognition}, pages 20083--20093, 2023.

\bibitem[Standley et~al.(2020)Standley, Zamir, Chen, Guibas, Malik, and Savarese]{standley2020tasks}
Trevor Standley, Amir Zamir, Dawn Chen, Leonidas Guibas, Jitendra Malik, and Silvio Savarese.
\newblock Which tasks should be learned together in multi-task learning?
\newblock In \emph{International Conference on Machine Learning}, pages 9120--9132. PMLR, 2020.

\bibitem[Suresh(2006)]{suresh2006mechanical}
Subra Suresh.
\newblock Mechanical response of human red blood cells in health and disease: Some structure-property-function relationships.
\newblock \emph{Journal of materials research}, 21\penalty0 (8):\penalty0 1871--1877, 2006.

\bibitem[Tang et~al.(2020)Tang, Liu, Zhao, and Gong]{tang2020progressive}
Hongyan Tang, Junning Liu, Ming Zhao, and Xudong Gong.
\newblock Progressive layered extraction (ple): A novel multi-task learning (mtl) model for personalized recommendations.
\newblock In \emph{Proceedings of the 14th ACM Conference on Recommender Systems}, pages 269--278, 2020.

\bibitem[Terven et~al.(2023)Terven, C{\'o}rdova-Esparza, and Romero-Gonz{\'a}lez]{terven2023comprehensive}
Juan Terven, Diana-Margarita C{\'o}rdova-Esparza, and Julio-Alejandro Romero-Gonz{\'a}lez.
\newblock A comprehensive review of yolo architectures in computer vision: From yolov1 to yolov8 and yolo-nas.
\newblock \emph{Machine Learning and Knowledge Extraction}, 5\penalty0 (4):\penalty0 1680--1716, 2023.

\bibitem[van Wyk and Bosman(2019)]{van2019evolutionary}
Gerard~Jacques van Wyk and Anna~Sergeevna Bosman.
\newblock Evolutionary neural architecture search for image restoration.
\newblock In \emph{2019 International Joint Conference on Neural Networks (IJCNN)}, pages 1--8. IEEE, 2019.

\bibitem[Wang et~al.(2020)Wang, Tsvetkov, Firat, and Cao]{wang2020gradient}
Zirui Wang, Yulia Tsvetkov, Orhan Firat, and Yuan Cao.
\newblock Gradient vaccine: Investigating and improving multi-task optimization in massively multilingual models.
\newblock \emph{arXiv preprint arXiv:2010.05874}, 2020.

\bibitem[Yang et~al.(2009)Yang, Kim, and Xing]{yang2009heterogeneous}
Xiaolin Yang, Seyoung Kim, and Eric Xing.
\newblock Heterogeneous multitask learning with joint sparsity constraints.
\newblock \emph{Advances in neural information processing systems}, 22, 2009.

\bibitem[Yu et~al.(2020)Yu, Kumar, Gupta, Levine, Hausman, and Finn]{yu2020gradient}
Tianhe Yu, Saurabh Kumar, Abhishek Gupta, Sergey Levine, Karol Hausman, and Chelsea Finn.
\newblock Gradient surgery for multi-task learning.
\newblock \emph{Advances in Neural Information Processing Systems}, 33:\penalty0 5824--5836, 2020.

\bibitem[Zhang and Yang(2021)]{zhang2021survey}
Yu Zhang and Qiang Yang.
\newblock A survey on multi-task learning.
\newblock \emph{IEEE Transactions on Knowledge and Data Engineering}, 34\penalty0 (12):\penalty0 5586--5609, 2021.

\bibitem[Zhang et~al.(2014)Zhang, Luo, Loy, and Tang]{zhang2014facial}
Zhanpeng Zhang, Ping Luo, Chen~Change Loy, and Xiaoou Tang.
\newblock Facial landmark detection by deep multi-task learning.
\newblock In \emph{Computer Vision--ECCV 2014: 13th European Conference, Zurich, Switzerland, September 6-12, 2014, Proceedings, Part VI 13}, pages 94--108. Springer, 2014.

\end{thebibliography}
